\documentclass[11pt,a4paper]{article}
\usepackage[hyperref]{ranlp2023}
\usepackage{times}
\usepackage{latexsym}
\usepackage{graphicx}
\usepackage{booktabs}

\usepackage{microtype}

\aclfinalcopy 

\title{Noisy Self-Training with Data Augmentations for Offensive and Hate Speech Detection Tasks}

\author{João A. Leite\textsuperscript{1} \and Carolina Scarton\textsuperscript{1} \and Diego F. Silva\textsuperscript{2} \\ \\
\textsuperscript{1}Department of Computer Science, The University of Sheffield, Sheffield (UK) \\ 
\textsuperscript{2}Instituto de Ciências Matemáticas e de Computação, Universidade de São Paulo, São Carlos (Brazil)\\
\texttt{\{jaleite1, c.scarton\}@sheffield.ac.uk}, \texttt{diegofsilva@usp.br}}

\date{}

\begin{document}
\maketitle
\begin{abstract}
Online social media is rife with offensive and hateful comments, prompting the need for their automatic detection given the sheer amount of posts created every second. Creating high-quality human-labelled datasets for this task is difficult and costly, especially because non-offensive posts are significantly more frequent than offensive ones. However, unlabelled data is abundant, easier, and cheaper to obtain. In this scenario, self-training methods, using weakly-labelled examples to increase the amount of training data, can be employed. Recent ``noisy'' self-training approaches incorporate data augmentation techniques to ensure prediction consistency and increase robustness against noisy data and adversarial attacks. In this paper, we experiment with default and noisy self-training using three different textual data augmentation techniques across five different pre-trained BERT architectures varying in size. We evaluate our experiments on two offensive/hate-speech datasets and demonstrate that (i) self-training consistently improves performance regardless of model size, resulting in up to +1.5\% F1-macro on both datasets, and (ii) noisy self-training with textual data augmentations, despite being successfully applied in similar settings, decreases performance on offensive and hate-speech domains when compared to the default method, even with state-of-the-art augmentations such as backtranslation. 

\end{abstract}

\section{Introduction}
Online social media platforms are widely used by modern society for many productive purposes. However, they are also known for intensifying offensive and hateful comments, attributed in part to factors such as user anonymity \cite{10.1145/3078714.3078723}. Manual identification of hate speech is impractical at scale due to the massive number of posts generated every second and the potential harm to the mental health of moderators. Therefore, there is a need for automatic approaches to detect offensive and hateful speech.

In recent years, research on this topic has increased, resulting in new models and datasets published in various languages and sources \cite{fortunahatespeechsurvey}. A common characteristic among available datasets is label skewness towards the negative class (non-offensive/hateful), which is usually more frequent than the positive class (offensive/hateful).
Apart from traditional ways of dealing with imbalanced classes (e.g. 
under or oversampling 
or applying class weighting), semi-supervised techniques such as self-training can be used to extend the training set with unseen examples that introduce new learning signals without the costly burden of manual data labeling.

Self-training is a technique that involves iteratively training models using both labelled and unlabelled data. The process begins by training a model using human-labelled data only, which is then used to infer labels for a set of unlabelled data, creating a weakly-labelled dataset. The weakly-labelled dataset and the human-labelled dataset are then aggregated and used to retrain the model. This iterative process is repeated for a fixed number of steps or until no performance improvement is observed. Self-training can be particularly useful when labelled data is scarce or expensive to obtain, and was successfully applied in a variety of domains such as computer vision \cite{10.1145/3577925}, audio and speech processing \cite{LIU2022100616}, and natural language processing \cite{selftrainingseq}. 

\begin{figure}[!htb]
\centering
\includegraphics{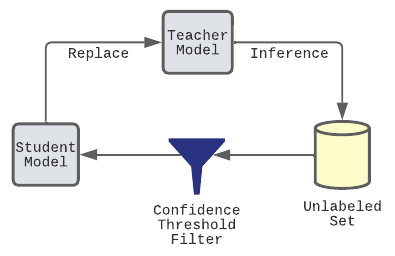}
\caption{Teacher-student self-training loop}
\label{fig:self-training-loop}
\end{figure}

Several variants of self-training have been proposed over the years \citep{amini2022self}. One common approach is to use a teacher-student framework, in which the ``student'' model learns from the output generated by the ``teacher'' model \citep{cotraining, noisystudent, crosspseudosupervision, selftrainingweaksupervision}. Additionally, a confidence threshold filter may be applied to remove examples that are too ambiguous or non-informative. This process is summarised in Figure \ref{fig:self-training-loop}. 

Recent research on self-training has reported further improvements in performance by introducing perturbations directly into the raw input or to its latent representation, improving generalisation and convergence \cite{10.5555/2969442.2969635, temporal_ensembling, miyato2018virtual, selftrainingseq, xie2020unsupervised}. These perturbations are often introduced in the form of data augmentations, which are widely applied in Computer Vision tasks but are less commonly explored in Natural Language Processing tasks, especially in the context of self-training. These ``noisy self-training'' methods can be particularly useful in settings where the input data is noisy or subject to a high degree of variation, improving prediction consistency and adversarial robustness \citep{carmon2019, wallach2019, najafi2019}.

\citet{bayer2022survey} argue that data augmentation depends on the underlying classification task, thus it cannot be effectively applied in all circumstances. Previous work focusing solely on data augmentation methods, not coupled with self-training, has shown mixed results for the domain of offensive/hate speech classification 
(Section \ref{ref:related_work_data_aug}). This indicates that there may not be a best method, while some may even negatively impact performance.

An open question is whether noisy self-training with text data augmentations can contribute to text classification tasks using state-of-the-art transfer-learning BERT models that have been shown to be invariant to various data transformations \citep{longpre-etal-2020-effective}. The task of offensive/abusive speech detection poses a difficult challenge for generating high-quality semantic invariant augmented examples, since it is a domain that is intrinsically associated with specific keywords that, if modified, can completely change the semantics of the text. In this paper, we innovate by providing an extensive experimentation setup using three different data augmentation techniques - backtranslation, random word swap, and random synonym substitution - in a self-training framework, with five different pre-trained BERT architectures varying in size, on two different datasets. 

We demonstrate that self-training, either with or without data noising, outperforms default fine-tuning regardless of model size, on both datasets. However, when comparing self-training without data noising vs `noisy' self-training, we find that data augmentations decrease performance, despite the literature reporting the superiority of noisy self-training in other domains. We further investigate how the augmentation methods fail to create label-invariant examples for the offensive/hate speech domain. Finally, we discuss future research ideas to address the limitations found in this work.

\section{Related Work}

\subsection{Data Augmentation} \label{ref:related_work_data_aug}
\citet{bayer2022survey} present a survey on data augmentation methods for NLP applications, reporting performance gains on various tasks.

In the domain of offensive/hate speech classification, \citet{imbalancedtoxic} experiment with three different text augmentation techniques to expand and balance their Wikipedia dataset by augmenting negative (non-offensive) examples. From a binary view of the dataset, more than 85\% of their examples are labelled as non-offensive, and from a multi-label view of the dataset, three of the six offensive classes are represented by less than 7\% of the dataset. They report F1-score increases of +1.4\% with unique words augmentation, +2.9\% with unique words and random mask, and +3.6\% with unique words, random mask, and synonym replacement.

\citet{mosolova2018text} use a custom synonym replacement augmentation method to experiment with a `toxic' dataset with 6 classes from a Kaggle competition\footnote{\url{https://www.kaggle.com/c/jigsaw-toxic-comment-classification-challenge}}. They experiment with character and word embeddings with a CNN architecture, and report a +3.7\% and +5.1\% ROC-AUC increase when applying their augmentation method with character embeddings on the public and private scores\footnote{Public scores are computed over a smaller portion of the test set. At the end of the competition, private scores are computed with the remainder of the test set.}, respectively. However, when coupled with word embeddings, they find that their augmentations result in a decrease of -0.09\% and -0.21\% ROC-AUC scores on the public and private scores, respectively.

\citet{10.1145/3357384.3358040} propose three text-based data augmentation techniques to address the class imbalance in datasets, and apply them on three English hate speech datasets named HON \cite{davidson2017automated}, RSN-1 \cite{waseemhovyhate} and RSN-2 \cite{waseem-2016-racist}. Their augmentation methods include (i) synonym replacement based on word embedding, (ii) warping of the token words along the padded sequence, and (iii) class-conditional RNN language generation. They compare the three methods on different architectures combining word embeddings, CNNs, GRUs, and LSTMs, and they report an average across four different architecture configurations of -6.3\% F1-Macro using (i), +5\% F1-Macro using (ii), and -4\% F1-Macro using (iii).

\citet{marivate2020improving} experiment with four different data augmentation techniques: WordNet synonym substitution, backtranslation between German and English, word embedding substitution according to cosine similarity, and mixup \cite{zhang2018mixup}.
Authors experiment with three datasets from different domains: Sentiment 140 \cite{go2009twitter}, AG News \cite{NIPS2015_250cf8b5} and a Hate Speech dataset \cite{davidson2017automated}. They observe performance increases on both Sentiment 140 and AG News across different augmentation methods, up to +0.4\% and +0.5\% accuracy score on AG News and Sentiment 140, respectively. However, they report performance decreases with all methods on the Hate Speech dataset, with decreases of 0.0\% with mixup, -0.3\% with embedding similarity, -0.8\% with synonym substitution, and -2.3\% with backtranslation. 

\subsection{Self-Training}

\citet{noisystudent} present a method called \textit{noisy student}, which achieves state-of-the-art results on the ImageNet dataset \citep{imagenet} by performing self-training with a teacher-student approach, using student models that are equal or larger-sized than the teacher models, and adding noise both to the input data through random image augmentations and to the model via dropout.

\citet{selftrainingseq} apply a similar idea using textual data augmentation methods such as backtranslation \citep{backtranslation} and token modifications to a self-training LSTM architecture for the tasks of machine translation and text summarization. They find that both model noise, in the form of dropout, and data noise, in the form of data augmentations, are crucial to their observed increase in performance on both tasks.

\citet{xie2020unsupervised} use six text classification and two image classification benchmark datasets to experiment with different types of noise-inducing techniques 
for self-training. They argue that state-of-the-art augmentations like backtranslation for text classification and RandAugment \citep{randaugment} for image classification, outperform simple noise inducing techniques, such as additive Gaussian noise.

The use of noisy self-training approaches in the domain of offensive/hate speech classification is still limited, but default `non-noisy' self-training has been successfully applied in some recent works. \citet{doi:10.1080/08839514.2021.1988443} collect unlabelled Arabic tweets and perform semi-supervised classification with self-training for the domain of Offensive and Hate Speech detection using multiple text representations such as N-grams, Word2Vec, AraBert and Distilbert, and multiple model architectures such as SVM, CNN and BiLSTM. They report up to 7\% performance increase in low resource settings where only a few labelled examples are available.


\citet{leonardelli2020dh} apply self-training in their submission to the HaSpeeDe shared task on Italian hate speech detection (task A). They fine-tune an AlBERTo model with the human-labelled dataset provided by the task organisers and extend it with a weakly-labelled dataset using self-training. Additionally, they oversample the human-labelled set in an attempt to make the model more robust to inconsistencies in the weakly-labelled set. Their submission achieve an F1-macro score of 75.3\% on tweets, placing 11th out of 29 teams, and 70.2\% on news headlines, placing 5th out of 29 teams.

\citet{pham-hong-chokshi-2020-pgsg} report experiments with the noisy student method from \citet{noisystudent} in the OffensEval 2020 shared task, achieving 2nd place at subtask B (Automatic categorization of offense types).
In their setup, although dropout is applied to a BERT-large model, no noise is injected into the data, which is a crucial component of the noisy student method. Because of this, we argue that this work is actually applying a default self-training method instead of a noisy self-training method. Also, OffensEval 2020's training data does not contain human-labelled data\footnote{In OffensEval 2020, the labels in the training data are the average confidence score and confidence standard deviation aggregated from an ensemble of models.}, thus both their weakly-labelled dataset and ground-truth dataset consist of inferred examples.

\citet{richardson2022semi} detect hate speech on Twitter in the context of the Covid-19 pandemic. They employ a simple approach, utilizing a bag-of-words representation combined with an SVM classifier. Authors demonstrate that by employing self-training with only 20\% of the training data, they manage to improve accuracy by +1.55\% compared to default training using 80\% of the training data.

To the best of our knowledge, \citet{santos_et_al:OASIcs.SLATE.2022.11} is the only previous work in which a \textbf{noisy} self-training approach was attempted on an offensive/hate speech classification task. They propose an ensemble of two semi-supervised models to create FIGHT, a Portuguese hate speech corpus. Authors combine GANs, a BERT-based model, and a label propagation model, achieving 66.4\% F1-score. They attempt to increase performance using backtranslation as data augmentation, but ultimately observe no performance gains, thus their best model is obtained with default self-training, not with noisy self-training. 


\section{Materials and Methods} \label{sec:experimental-setup}
This section presents the description of the datasets, data augmentation methods and self-training architectures used throughout our experiments. Our code is available at GitHub\footnote{\url{https://github.com/JAugusto97/Offense-Self-Training}}.


\subsection{Data Description} \label{sec:data-description}
We use two English binary offensive/hate speech detection datasets in our experiments. Table~\ref{tab:datasets-distributions} presents their target class distributions. 

\begin{table}[htb]
\centering
\begin{tabular}{lccc}
\hline
\multicolumn{4}{c}{OLID}                                       \\ 
         & Train           & Dev             & Test            \\ \hline
Not-Offensive & 8,840  & 0               & 620   \\
Offensive & 4,400  & 0             & 240   \\ \hline
\multicolumn{4}{c}{ConvAbuse}                                  \\ 
         & Train           & Dev             & Test            \\ \hline
Not-Offensive & 2,163  & 719   & 725   \\
Offensive & 338   & 112   & 128   \\ \hline
\end{tabular}%
\caption{\label{tab:datasets-distributions}Target class distribution for OLID and ConvAbuse.}
\end{table}


\paragraph{Offensive Language Identification Dataset (OLID)} \citep{oliddataset} contains a collection of annotated tweets following three levels: Offensive Language Detection, Categorization of Offensive Language, and Offensive Language Target Identification. This work only uses the first level - Offensive Language Detection. The dataset was normalised by replacing URLs and user mentions with placeholders. The best model in \citep{oliddataset} achieves 80\% macro-$F1$ using convolutional neural networks, with 70\% and 90\% of $F1$-Score for the positive and negative classes, respectively.


\paragraph{ConvAbuse} \citep{convabusedataset} is a dataset on abusive language towards three conversational AI systems: an open-domain social bot, a rule-based chatbot, and a task-based system. Authors find that the distribution of abuse towards conversational systems differs from other commonly used datasets, with more than 50\% of the instances containing sexism or sexual harassment. To normalise the data, web addresses were replaced with a placeholder. Authors provide standard train, development, and test sets and achieve up to 88.92\% macro-$F1$ using a fine-tuned BERT model. In our experiments, we concatenate the interactions between the user and the chatbot into a single text document divided by new line separators, and we use majority voting between the annotations to consolidate the binary abusive vs. non-abusive label.

\paragraph{Unlabelled data} We collected 365,456 tweets in English with the Twitter API using an unbiased query rule: random tweets mentioning stop-words like ``in'', ``on'', ``a'', ``is'', ``not'', ``or'' and so on. We also preprocess the data by removing user mentions, urls, punctuations, extra whitespace and accents.

\subsection{Self-Training Architecture} \label{sec:system-description}
Our noisy self-training system is similar to that introduced by \citet{noisystudent} and \citet{xie2020unsupervised}, and works as follows:

\begin{enumerate}
\item A teacher model is trained to minimise the cross-entropy loss on the human-labelled training set exclusively.
\item The teacher model infers weak labels from the unlabelled dataset. \label{steptwo}
\begin{itemize}
\item A confidence threshold filter is applied, and examples that fall below this threshold are removed.
\item Apply \textit{downsampling} on the inferred examples, ending up with a perfectly balanced weakly-labelled dataset.
\end{itemize}
\item All the examples selected from the previous step are augmented once with one of the data augmentation methods, doubling the amount of weakly-labelled examples. The labels obtained with the `clean/without noise' text in step \ref{steptwo} are replicated for the augmented texts. \label{stepthree}
\item An equal-sized student model minimises the combined cross-entropy loss on human-labelled and weakly-labelled datasets:
\begin{equation}
L =\frac{1}{n}\sum_{i=1}^{n}L_{\mathrm{labelled}}+\frac{1}{m}\sum_{i=1}^{m}L_{\mathrm{inferred}}
\label{eq:loss}
\end{equation}
\item Repeat from step \ref{steptwo} using the current student model as the teacher model.

\end{enumerate}

In our experiments, we compare this noisy self-training framework against the default 'non-noisy' self-training method, which simply skips step \ref{stepthree}, meaning we do not apply any form of data augmentation.

\subsection{Data Augmentation Methods} \label{sec:data_augmentations}
In each noisy self-training experiment we use \texttt{nlpaug}\footnote{\url{https://github.com/makcedward/nlpaug}} to apply one of the three following data augmentation methods for textual data:

\paragraph*{Random Synonym Substitution} Uses WordNet \cite{miller1995wordnet} to randomly replace tokens by one of its synonyms. For each sentence, 30\% of its tokens will be replaced.
\paragraph*{Random Word Swap} Randomly swaps adjacent tokens in a sentence. For each sentence, 30\% of its tokens are swapped.
\paragraph*{Backtranslation} First translates the original texts into a second language, then translates them back from the second language to the original language. We use the backtranslation model from \texttt{nlpaug}, which uses
the two different transformer models from \citet{ng-etal-2019-facebook} to translate the data from English to German, then from German back to English.

\section{Experimental Setup}
Firstly, we experiment with each dataset to estimate the hyperparameters for the base models, which is the first teacher models in the self-training loop. We use a batch size of 128, maximum sequence length of 128, learning rate of 0.00001, 15\% of the training set as warm-up batches, weight decay of 0.001 and 20 training epochs. We apply a dropout rate of 10\% for both the attention and classification layers. The model with highest validation F1-macro score\footnote{Lowest training loss in the case of OLID, since no development set is provided.} obtained during training is loaded at the end of the last epoch. For the hyperparameters associated with the self-training method, we set the number of teacher-student iterations to 4 (including the first teacher model) and a confidence threshold filter of 80\%, similarly to \citet{xie2020unsupervised}. Also, we experiment with five different pre-trained BERT models: DistilBERT, BERT-base-cased, BERT-large-cased, RoBERTa-base and RoBERTa-large, aiming to investigate the impact of model size in performance gains associated with self-training.

From the above-listed configurations, we designed two main classification scenarios. The first scenario accounts for a regular self-training loop without data noise injection through augmentations, while the second scenario uses the noisy self-training approach, introducing data noise with one of the three augmentation methods described in Section \ref{sec:data_augmentations}.

Finally, we conduct a deeper analysis of each augmentation method. We use the first teacher model, trained exclusively with the human-labelled data of each dataset, to infer both the 'clean/without augmentation' and the 'noisy/augmented' versions of the unlabelled dataset and verify the following: (i) Does the augmentation method create new tokens that are not present in the vocabulary of the 'clean/without augmentation' unlabelled dataset? and (ii) Are the augmentations semantically invariant, meaning both the 'clean' and 'noisy' pairs of examples are assigned the same label?

\section{Results}

\begin{table*}
\centering
\begin{tabular}{@{}lccccc@{}}
\toprule
\multicolumn{6}{c}{OLID}                                                                                      \\
Architecture     & DF         & ST                  & ST + BT     & ST + SS             & ST + WS             \\ \midrule
DistilBERT  & 78.4\footnotesize{ ± 0.1} & \textbf{79.2\footnotesize{ ± 0.2}} & 79.0\footnotesize{ ± 0.3}  & 79.0\footnotesize{ ± 0.3}          & 79.0\footnotesize{ ± 0.3}          \\
BERT-base-cased  & 77.2\footnotesize{ ± 0.3} & \textbf{78.7\footnotesize{ ± 0.1}} & 78.1\footnotesize{ ± 0.1}  & 78.3\footnotesize{ ± 0.3}          & 78.3\footnotesize{ ± 0.3}          \\
BERT-large-cased & 79.2\footnotesize{ ± 0.2} & \textbf{80.0\footnotesize{ ± 0.3}} & 79.4\footnotesize{ ± 0.1}  & 79.3\footnotesize{ ± 0.3}          & 79.3\footnotesize{ ± 0.3}          \\
RoBERTa-base     & 79.4\footnotesize{ ± 0.7} & \textbf{80.1\footnotesize{ ± 0.3}} & 80.0\footnotesize{ ± 0.4}  & 80.0\footnotesize{ ± 0.4}          & 80.0\footnotesize{ ± 0.4}          \\
RoBERTa-large    & 79.8\footnotesize{ ± 0.3} & 80.4\footnotesize{ ± 0.4}          & 80.3\footnotesize{ ± 0.4} & \textbf{80.7\footnotesize{ ± 0.7}} & \textbf{80.7\footnotesize{ ± 0.7}} \\ \midrule
\multicolumn{6}{c}{ConvAbuse}                                                                                 \\
Architecture     & DF         & ST                  & ST + BT     & ST + SS             & ST + WS             \\ \midrule
DistilBERT  & 85.7\footnotesize{ ± 0.5} & 86.8\footnotesize{ ± 0.3}          & 87.1\footnotesize{ ± 0.3}  & \textbf{87.2\footnotesize{ ± 0.3}} & \textbf{87.2\footnotesize{ ± 0.3}} \\
BERT-base-cased  & 86.8\footnotesize{ ± 0.8} & \textbf{87.6\footnotesize{ ± 0.1}} & 87.2\footnotesize{ ± 0.5}  & 87.2\footnotesize{ ± 0.5}          & 87.2\footnotesize{ ± 0.5}          \\
BERT-large-cased & 87.1\footnotesize{ ± 0.6} & \textbf{87.9\footnotesize{ ± 0.5}} & 87.4\footnotesize{ ± 0.2}  & \textbf{87.9\footnotesize{ ± 0.5}} & \textbf{87.9\footnotesize{ ± 0.5}} \\
RoBERTa-base     & 84.5\footnotesize{ ± 0.3} & \textbf{85.5\footnotesize{ ± 0.4}} & 85.3\footnotesize{ ± 0.8}  & 85.4\footnotesize{ ± 0.5}          & 85.4\footnotesize{ ± 0.5}          \\
RoBERTa-large    & 86.0\footnotesize{ ± 0.1} & 86.2\footnotesize{ ± 0.3}          & 86.6\footnotesize{ ± 0.3} & \textbf{86.9\footnotesize{ ± 0.1}} & 86.8\footnotesize{ ± 0.1}         \\ \bottomrule
\end{tabular}
\caption{Mean ± 1 std F1-Macro scores obtained over three random seed initializations.\\ \centering DF=Default Fine-Tuning, ST=Self-Training, BT=Backtranslation, SS=Synonym Substitution, WS=Word Swap}
\label{tab:results}
\end{table*}

\subsection{Default Fine-Tuning vs. Self-Training} \label{sec:dfvsst}
Table \ref{tab:results} displays the mean and standard deviation $F1$-macro scores computed over three different random seed initializations for each experiment. Note that self-training, regardless of whether coupled with data augmentation methods or not, improves over default fine-tuning for every model architecture, increasing the F1-macro score from +0.7\% up to +1.5\% on OLID and +0.8\% up to +1.5\% on ConvAbuse depending on the pre-trained model architecture.

Also, we highlight how self-training can make smaller models, which require fewer resources to maintain in practical applications, achieving the same performance as larger and more costly models that are trained with default fine-tuning. Self-training on a DistilBERT (66M parameters) outperforms a BERT-large-cased (340M parameters) with default fine-tuning on both OLID and ConvAbuse. On OLID, a RoBERTa-base architecture (125M parameters) with self-training outperforms a RoBERTa-large (354M parameters) architecture with default fine-tuning, although this does not hold true for ConvAbuse.

Furthermore, we point out that OLID and ConvAbuse's data come from different sources, the first being Twitter, and the second one representing conversations between humans and chatbots, thus their structure differs significantly. Since our unlabelled dataset is composed of Twitter data, it would be fair to assume that the benefits of self-training in our experiments would be more prominent for the OLID dataset, but our results do not show this, since models trained with ConvAbuse benefited from self-training with our Twitter-originated unlabelled dataset just as much as models trained with OLID.

\subsection{Default Self-Training vs. Noisy Self-Training}
After verifying that self-training is beneficial to both datasets on all model architectures, we compare default self-training with noisy self-training, and the impacts of adding data noise in the form of data augmentations. We find that introducing data augmentations to the self-training pipeline increases performance against default self-training only for RoBERTa-large on both OLID and ConvAbuse, with DistilBERT also showing improvements for ConvAbuse, but not for OLID. On all other architectures, for both datasets, default self-training without data augmentations achieves the highest scores. 

In our results for offensive/hate speech classification, backtranslation does not achieve the highest score in any setup, while synonym substitution and word swap tie for highest score in three scenarios: ConvAbuse with DistilBERT, ConvAbuse with BERT-large-cased, and OLID with RoBERTa-large. Synonym substitution outperforms all the remaining methods on ConvAbuse with RoBERTa-large.

An important remark is that our results diverge from \citet{selftrainingseq}, which finds that state-of-the-art data augmentation methods such as backtranslation outperform simpler methods on self-training for machine translation and text summarization. However, our results align with \citet{marivate2020improving}, although their work is not focused on self-training, but instead on how different data augmentation techniques impact their models on three datasets from different domains. They report backtranslation as their worst augmentation method on a hate speech dataset, decreasing accuracy by -2.3\%. Our findings bridge this gap and reveal that backtranslation has significant limitations in the domain of offensive/hate speech detection, even when used in a noisy self-training approach.

\subsection{Data Augmentation Analysis}
Our first data augmentation analysis is to understand if the augmented text introduces new unseen tokens to the vocabulary of the `clean' unlabelled set when both are combined. We find a vocabulary size increase of 39.5\%, 9.0\% and 4.7\% averaging across all different pre-trained architectures for backtranslation, synonym substitution and word swap\footnote{Word swap is unintuitively capable of creating new tokens depending on how a sentence is split into tokens and then merged back after swapping the tokens.} respectively. This indicates that backtranslation is heavily superior in terms of introducing new unseen tokens, but this is not correlated with performance increase, as backtranslation appears as the worst augmentation method for noisy self-training in our classification experiments.

Next, in order to verify the performance of the data augmentation methods in generating semantically invariant examples, we use the base models trained exclusively with the human-labelled data from each dataset, on each pre-trained architecture, and use them to perform inference on both the `clean' and the noisy/augmented unlabelled set. We then compare both predictions and analyse how augmentations may shift the underlying target class. We will refer to \textbf{positive shift} when a non-offensive example is classified as offensive after being augmented, and \textbf{negative shift} when an offensive example is classified as non-offensive after being augmented.

Table \ref{tab:label-shift} presents the total class shift percentage for each augmentation method, averaging across both datasets and all model architectures, of which we further divide into positive and negative label shift percentages. Notice that backtranslation is the method that produces the highest amount of label shifting at 23.8\%, of which 54.7\% are negative shifts, which is a 6.6\% increase over synonym substitution and a 4.8\% increase word swap.

\begin{table}[htb]
\centering
\resizebox{\columnwidth}{!}{%
\begin{tabular}{@{}lc|cc@{}}
\toprule
Augmentation & \multicolumn{1}{l|}{Total Shift} & \multicolumn{1}{l}{Positive Shift} & \multicolumn{1}{l}{Negative Shift} \\ \midrule
BT           & 23.8\%                           & 46.7\%                             & 54.7\%                             \\
SS           & 23.5\%                           & 48.7\%                             & 51.3\%                             \\
WS           & 23.3\%                           & 47.8\%                             & 52.2\%                             \\ \bottomrule
\end{tabular}
}
\caption{\label{tab:label-shift}Average target class shift percentage on the weakly-labelled set. BT=Backtranslation, SS=Synonym Substitution, WS=Word Swap}
\end{table}

It is fair to assume that not all of the class shifting occurs from the augmentation changing the semantic that defines if an example is either offensive or not-offensive. In most cases, class shifting may occur because of small perturbations that are semantically invariant, meaning both the 'clean' and the augmented text's true underlying classes are still the same, even if the classifier predicted them as different classes. In these cases, when we set the label of the augmented text to be the same as the one obtained when inferring the 'clean' version of the text, as presented in section \ref{sec:system-description}, we are reinforcing the model to be more robust against these small perturbations, which is one of the main benefits of noisy self-training. However, when augmentation methods create semantically different versions of the original texts, replicating the inferred label from the original text to the augmented text results in the addition of incorrect ground-truth labels to the train set, which may degrade performance.

Currently, to the best of our knowledge, there is no dataset annotated for offense/hate speech before and after applying data augmentation, which would enable a more accurate estimation of semantic variations produced by them. In tables \ref{tab:positive-shift} and \ref{tab:negative-shift} we show two examples for each augmentation method that suffered from positive shift (not-offensive to offensive) and negative shift (offensive to not-offensive), respectively.

\begin{table*}[]
\centering
\begin{tabular}{lll}
\hline
Text                         & Augmented Text                      & Method \\ \hline
I HATE ALL OF YOU & ALL I HATE OF YOU & WS \\
Maybe I dont respect all women                   & Maybe I respect dont women all                      & WS                         \\
Bitches and sports                               & Females and Sport                                   & BT                         \\
Wooooow what the fuck                            & Wooooow, what the hell?                             & BT                         \\
Bitch you better be joking                       & Gripe you good be joking                            & SS                         \\
The NYT has been showing its whole ass {[}...{]} & The NYT has follow showing its whole butt {[}...{]} & SS                         \\ \hline
\end{tabular}

\caption{\centering Examples of Offensive to Not-Offensive semantic shift created by data augmentation. \newline BT=Backtranslation, SS=Synonym Substitution, WS=Word Swap}
\label{tab:positive-shift}
\end{table*}

\begin{table*}[]
\centering
\begin{tabular}{lll}
\hline
Text                        & Augmented Text                         & Method \\ \hline
Is that Fat Albert & That Fat is Albert & WS \\
Man that is terrible & That man is terrible & WS \\
damn white people oppressing the blacks   & fucking white people who oppress the blacks      & BT                         \\
That damn staircase be beating my ass {[}...{]} & That fucking staircase will bang my ass {[}...{]}      & BT                         \\
i will not get over this                        & i will not fuck off ended this                         & SS                         \\
Send me the link and Ill love you forever      & Send pine tree state the link and Ill fuck you forever & SS                         \\ \hline
\end{tabular}
\caption{\centering Examples of Not-Offensive to Offensive class shift created by data augmentation. \newline BT=Backtranslation, SS=Synonym Substitution, WS=Word Swap}
\label{tab:negative-shift}
\end{table*}

An example of a recurrent theme among various target shifted examples is the substitution of the keywords `fuck' with `damn' or `hell', indicating that despite these keywords being semantically similar, they are not always interchangeable with respect to the target class, and the mere replacement of one for another is enough to shift the target class. This could be expected, as offense detection is highly impacted by the mere presence or absence of offensive keywords.

\section{Conclusion}
In this work, we analysed the impact of self-training on offensive and hate speech classification tasks using five different pre-trained BERT models of varying sizes and two different datasets. We also experimented with noisy self-training using three different data augmentation techniques for textual data. We found that self-training improves classification performance for all model architectures on both datasets, with an increase in F1-Macro of up to +1.5\%. However, our experiments comparing default self-training versus noisy self-training showed that noisy self-training does not improve performance, despite its success in other domains. Finally, we investigated the three data augmentation methods and showed that the domain of offensive/hate speech classification is highly sensitive to semantic variances produced by them, and we discussed future research ideas to mitigate these problems.

\section{Future Work}


We understand that some of the semantic variations discussed in this work could be mitigated by data augmentation methods that both preserve existing offensive keywords, and do not introduce new offensive keywords randomly, as these are often conditional to the underlying ground-truth class. For some languages, most of these keywords are extensively documented\footnote{\url{https://hatebase.org/}}, thus they can be known a priori by these methods, and be treated differently, such as only substituting an offensive keyword by another offensive keyword, or not allowing a non-offensive keyword to be substituted by an offensive keyword. This custom approach can theoretically help mitigate semantic variations in this domain, but offensive/hateful comments can still be made without making use of a single offensive/hateful keyword. In these more subtle cases, a system would have to detect the offensive/hateful context without relying solely on keywords, and modify the example while still maintaining this context. We see potential benefits of using recent instruction-tuned large language models \cite{ouyang2022training} as specialised data augmentation methods that are task-specific, and can be able to preserve the semantics associated with the task when modifying a given text. In this scenario, an instruction prompt can be designed to inform the system of the context of the task, and make it aware that this semantic must be preserved when modifying the given text. In the future, we aim towards extending this work with the above-mentioned research ideas.

\section*{Acknowledgments}
We thank Olesya Razuvayevskaya and Freddy Heppell for their valuable feedback. This research has been funded by "SoBigData++: European Integrated Infrastructure for Social Mining and Big Data Analytics” (EU H2020, Grant Agreement n.871042 (\url{http://www.sobigdata.eu})).


\bibliographystyle{acl_natbib}
\bibliography{ranlp2023}


\end{document}